\documentclass[final]{cvpr}

\usepackage{times}
\usepackage{epsfig}
\usepackage{graphicx}
\usepackage{amsmath}
\usepackage{amssymb}

\usepackage{comment}
\usepackage{booktabs, caption, makecell}
\usepackage{epstopdf}
\usepackage{cite}
\usepackage{tabularx}
\usepackage[normalem]{ulem}

\usepackage{color}
\usepackage{subfig}
\usepackage{multirow}
\usepackage{array}
\usepackage{enumitem}
\usepackage{mathtools}

\usepackage[pagebackref=true,breaklinks=true,colorlinks,bookmarks=false]{hyperref}

\newcolumntype{L}[1]{>{\raggedright\let\newline\\\arraybackslash\hspace{0pt}}m{#1}}
\newcolumntype{C}[1]{>{\centering\let\newline\\\arraybackslash\hspace{0pt}}m{#1}}
\newcolumntype{R}[1]{>{\raggedleft\let\newline\\\arraybackslash\hspace{0pt}}m{#1}}
\newcolumntype{+}{>{\global\let\currentrowstyle\relax}}
\newcolumntype{^}{>{\currentrowstyle}}

\newcommand{\bfF}{\ensuremath{{\mathbf{F}}}}
\newcommand{\bfG}{\ensuremath{{\mathbf{G}}}}
\newcommand{\bfH}{\ensuremath{{\mathbf{H}}}}
\newcommand{\bfA}{\ensuremath{{\mathbf{A}}}}

\newcommand{\bfD}{\ensuremath{{\mathbf{D}}}}

\newcommand{\bfS}{\ensuremath{{\mathbf{S}}}}

\newcommand{\bfE}{\ensuremath{{\mathbf{E}}}}

\newcommand{\bfY}{\ensuremath{{\mathbf{Y}}}}

\newcommand{\calA}{\ensuremath{{\mathcal{A}}}}
\newcommand{\calE}{\ensuremath{{\mathcal{E}}}}

\newcommand{\calI}{\ensuremath{{\mathcal{I}}}}

\newcommand{\calF}{\ensuremath{{\mathcal{F}}}}

\newcommand{\calU}{\ensuremath{{\mathcal{U}}}}

\newcommand{\calO}{\ensuremath{{\mathcal{O}}}}

\def\cvprPaperID{4551} 
\def\confYear{CVPR 2021}

\begin{document}

\title{Guided Interactive Video Object Segmentation \\ Using Reliability-Based Attention Maps}

\author{Yuk Heo\\
Korea University\\
{\tt\small yukheo@mcl.korea.ac.kr}
\and
Yeong Jun Koh\\
Chungnam National University\\
{\tt\small yjkoh@cnu.ac.kr}
\and
Chang-Su Kim\\
Korea University\\
{\tt\small changsukim@korea.ac.kr}
}

\maketitle

\begin{abstract}
We propose a novel guided interactive segmentation (GIS) algorithm for video objects to improve the segmentation accuracy and reduce the interaction time. First, we design the reliability-based attention module to analyze the reliability of multiple annotated frames. Second, we develop the intersection-aware propagation module to propagate segmentation results to neighboring frames. Third, we introduce the GIS mechanism for a user to select unsatisfactory frames quickly with less effort. Experimental results demonstrate that the proposed algorithm provides more accurate segmentation results at a faster speed than conventional algorithms. Codes are available at \href{https://github.com/yuk6heo/GIS-RAmap}{https://github.com/yuk6heo/GIS-RAmap}.
\end{abstract}

\vspace{-0.1cm}
\section{Introduction}
\vspace{-0.1cm}

Video object segmentation (VOS) is a task to cut out objects of interest in a video. It is useful in various applications such as video editing, video summarization, video inpainting, and self-driving cars. VOS is challenging since it should deal with multiple objects, object deformation, and object occlusion. Because of this difficulty, semi-supervised VOS, which uses a fully annotated segmentation mask in the first frame, has been widely researched. This approach can improve the segmentation performance but requires a lot of time and effort for annotations (\eg around 79 seconds per instance~\cite{DAVISchallenge2018}). Also, it does not have a fallback mechanism when unsatisfactory results are obtained.

\begin{figure}[t]
\centering
    \includegraphics[width=\linewidth]{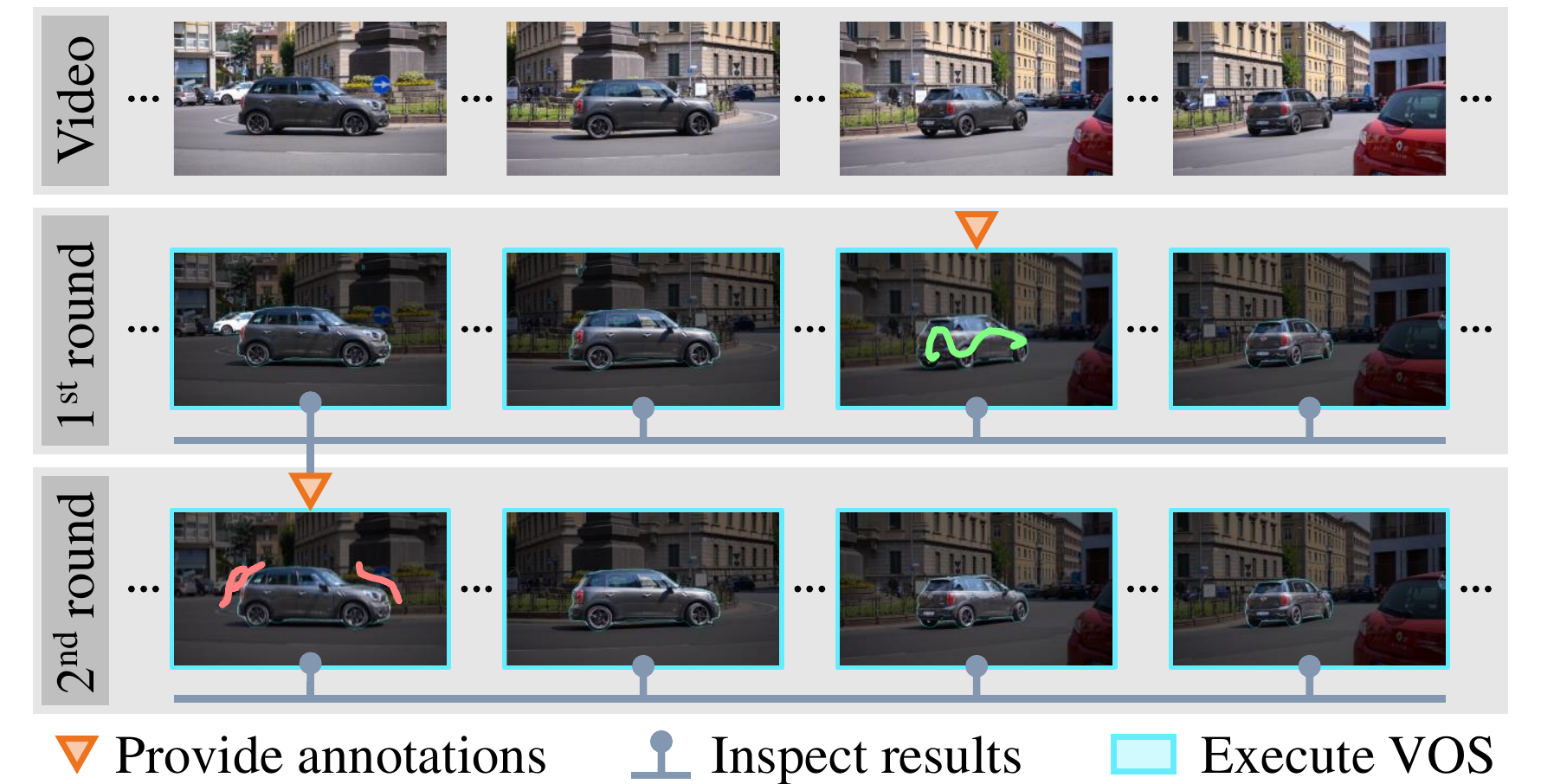}
\caption{Illustration of the round-based interactive VOS. The proposed algorithm greatly reduces the inspection time, by guiding users to select a frame for annotations efficiently and effectively. It is recommended to watch the supplementary video with a real-time demo of the proposed guided interactive system.}
\label{fig:VOSIntro}
\vspace{-0.3cm}
\end{figure}

Interactive VOS adopts user-friendly annotations, \eg scribbles, which are simple enough to provide repeatedly. Figure~\ref{fig:VOSIntro} shows the round-based interactive VOS process. First, a user selects a target frame and draws scribble annotations on it. After extracting query object information from the scribbles, the algorithm obtains segmentation results for all frames. Second, the user finds a frame with unsatisfactory results and then provides additional scribbles. The algorithm then exploits both sets of scribbles to refine the VOS results. This is repeated until the user is satisfied.

Recently, an automatic simulation scheme of the round-based interactive VOS was designed in~\cite{DAVISchallenge2018}. However, the simulation is significantly different from real applications in that it immediately determines segmentation results to correct, by comparing them with the ground-truth. In contrast, a real user should spend considerable time to inspect the results and select poorly segmented regions. Since conventional interactive VOS algorithms~\cite{Oh2019CVPR, Miao2020CVPR, Heo2020ECCV, Oh2020TPAMI} have been developed based on the simulation in~\cite{DAVISchallenge2018}, they do not consider the time for finding unsatisfactory results in practice. In contrast, we propose a guided interactive segmentation (GIS) algorithm for video objects, which guides users to find poorly segmented regions quickly and effectively.

Moreover, although interactive VOS can use the information in $N$ annotated frames in the $N$th round, the conventional algorithms~\cite{Miao2020CVPR, Heo2020ECCV} do not exploit those multiple annotated frames thoroughly. Heo \etal~\cite{Heo2020ECCV} simply average the features from multiple annotated frames. Miao \etal~\cite{Miao2020CVPR} use only the best matching result between a target frame and multiple annotated frames. On the contrary, we analyze the reliability of each annotated frame to refine segmentation results in a target frame more accurately.

In this paper, we propose the GIS algorithm using reliability-based attention (R-attention) maps. First, we transfer query object information from annotated frames to a target frame using R-attention maps, which represent pixel-wise reliability of the annotated frames. Next, we perform intersection-aware propagation to propagate segmentation results to neighboring frames sequentially. Third, we compute a guidance score, called R-score, to reduce or remove the processing time for selecting the frame to be annotated in each round. Experimental results demonstrate that the proposed GIS algorithm outperforms recent state-of-the-arts in both the interactive VOS simulation in~\cite{DAVISchallenge2018} and real-world applications.

This paper has three main contributions:\vspace{-0.1cm}
\begin{enumerate}
    \item Two novel operators to exploit multiple annotations and neighboring results are developed for VOS: R-attention and intersection-aware propagation modules.

    \vspace{-0.2cm}
    \item We propose the notion of guidance in interactive VOS.

    \vspace{-0.2cm}
    \item The proposed GIS algorithm outperforms the state-of-the-arts significantly in both speed and accuracy.
\end{enumerate}

\vspace{-0.2cm}
\section{Related Work}
\vspace{-0.1cm}

\noindent{\bf Unsupervised VOS:}
It is a task to find primary objects~\cite{Koh2016pod} without any user annotations in video sequences. Traditional approaches \cite{wang2015saliency, papazoglou2013fast, jang2016primary, Koh2017primary, Koh2018} use motion, object proposals, or saliency to solve this problem. Recently, with the availability of big VOS datasets \cite{DAVIS2017,Youtube2018}, many deep-learning-based unsupervised methods \cite{jain2017fusionseg,tokmakov2017learning,lu2019see,yang2019anchor,wang2019learning,zhou2020matnet,zhen2020learning} have been proposed.

\vspace*{0.15cm}
\noindent{\bf Semi-supervised VOS:}
In semi-supervised VOS, a user provides fully annotated masks for target objects at the first frame. Many algorithms have been developed to extract significant features for target objects using user annotations. Early deep learning methods \cite{caelles2017one,bao2018cnn,voigtlaender2017online,maninis2018video} focused on fine-tuning networks using annotation masks at the first frames. Instead of the computationally demanding fine-tuning, some algorithms employ optical flow for initial segment propagation \cite{Jang2017CVPR} or motion feature extraction \cite{hu2018motion}. Also, networks to refine segmentation results in previous frames without using motion have been developed in \cite{wug2018fast,lin2019agss}.
Instead of the first frame, optimal frames to be annotated in semi-supervised VOS were determined in \cite{griffin2019bubblenets}. Matching-based algorithms \cite{chen2018blazingly, hu2018videomatch, voigtlaender2019feelvos,Oh2019ICCV,Oh2020TPAMI} perform pixel-wise feature matching between annotated and target frames to segment out target objects. For example, key-value memory operations based on non-local networks \cite{wang2018non} are performed in \cite{Oh2020TPAMI, Oh2019ICCV} to perform the matching between a target frame and already segmented frames. In~\cite{wu2020memory}, the selection network, which predicts scores of previously segmented frames, is used to choose the frames for segmentation propagation.

\vspace*{0.15cm}
\noindent{\bf Interactive VOS:}
It aims at achieving satisfactory VOS results through an iterative process of drawing simple annotations, such as scribbles, point clicks, or bounding boxes. Early interactive VOS algorithms~\cite{wang2005interactive, price2009livecut, shankar2015video} constructed graph models, by connecting pixels with edges and then assigning edge weights using hand-craft features. Segmentation results for query objects were then obtained by graph optimization techniques.

Recently, deep-learning methods have been developed for interactive VOS. Benard and Gygli~\cite{benard2017interactive} used point clicks to extract an object mask in a single frame and apply a semi-supervised VOS algorithm to propagate the mask. Chen~\etal~\cite{chen2018blazingly} employed pixel-wise metric learning to cut out a query object using a few point clicks. Caelles~\etal~\cite{DAVISchallenge2018} introduced a round-based interactive VOS process and the automatic simulation algorithm to mimic human interactions in real applications. Many recent interactive algorithms~\cite{Oh2019CVPR, Miao2020CVPR, Heo2020ECCV, Oh2020TPAMI} follow this round-based process.

Oh~\etal~\cite{Oh2019CVPR} developed two segmentation networks for interactive VOS: the first one estimates target object regions from user interactions and the second one propagates the segmentation results to neighboring frames. Miao~\etal~\cite{Miao2020CVPR} proposed networks to obtain segmentation results through both interaction and propagation. They employed global and local distance maps in~\cite{voigtlaender2019feelvos} to match a target frame to an annotated frame and the previous frame, respectively. Heo~\etal~\cite{Heo2020ECCV} designed global and local transfer modules to effectively transfer features in annotated and previous frames to a target frame. Oh~\etal~\cite{Oh2020TPAMI} encoded annotation regions into keys and values in a non-local manner. These interactive VOS algorithms~\cite{Oh2019CVPR,Miao2020CVPR,Heo2020ECCV,Oh2020TPAMI}, however, have limitations. First, they do not consider the processing time to select the poorest segmentation results, on which additional annotations are provided. Second, they do not fully exploit the property that scribble data in multiple annotated frames have different reliability and different relevance to a target frame.

\begin{figure*}[t]
\centering
    \includegraphics[width=0.94\linewidth]{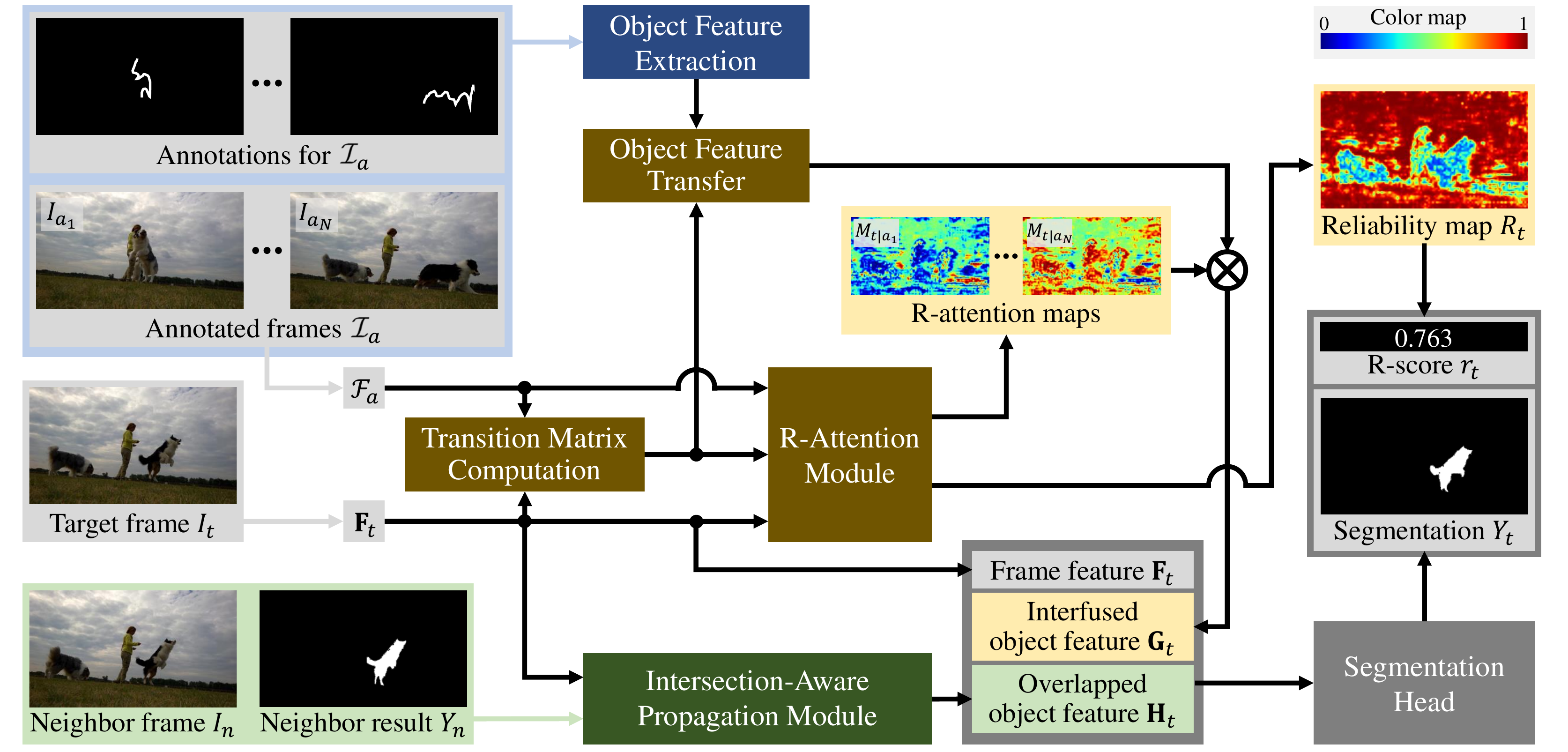}
\caption{An overview of the proposed GIS algorithm.}
\label{fig:OverallFlow}
\vspace{-0.3cm}
\end{figure*}

\vspace{-0.1cm}
\section{Proposed Algorithm}
\label{sec3:Proposed Algorithm}
\vspace{-0.1cm}

The proposed algorithm cuts query objects off in a video $\calI=\{I_1, \ldots, I_T\}$ with $T$ frames interactively using sparse annotations (scribbles or points) in each segmentation round. Let $I_{a_i}$ at time instance $a_i$ be the annotated frame in the $i$th round. First, the segmentation is performed bidirectionally starting from $I_{a_1}$. Subsequently, in the $N$-th round, it is also done bidirectionally, but using all previously annotated frames $\calI_a=\{I_{a_1}, \ldots, I_{a_N}\}$.

Figure~\ref{fig:OverallFlow} shows how the proposed algorithm segments a target frame $I_t$. First, we encode $I_t$ into the frame feature $\bfF_t$. Second, we obtain the interfused object feature $\bfG_{t}$ by combining query object information in all annotated frames in $\calI_a$ using the R-attention module. Third, using the neighbor frame $I_n\in\{I_{t-1}, I_{t+1}\}$ and its segmentation result $Y_n$, we perform the intersection-aware propagation to yield the overlapped object feature $\bfH_{t}$. 
Last, the segmentation head decodes these three features $\bfF_t$, $\bfG_t$, and $\bfH_t$ to generate the segmentation result ${Y}_{t}$ of the target frame $I_t$.

Moreover, we propose a novel guidance mechanism for interactive VOS. From the R-attention module, we extract the reliability map $R_t$ to represent pixel-wise reliability of the segmentation result $Y_t$. Also, by averaging these pixel-wise scores, we obtain the R-score $r_t$. These guidance data $R_t$ and $r_t$ enable a user to select a less reliable frame and provide annotations on it more quickly and more effectively.

\subsection{Interfused Object Feature}
\label{subsec:affinity}

As in \cite{Heo2020ECCV}, we transfer segmentation information in each annotated frame into a target frame. However, whereas \cite{Heo2020ECCV} combines the information from multiple annotated frames simply through averaging, we fuse transferred object features based on their reliability levels. To this end, we develop the R-attention module.

\vspace*{0.15cm}
\noindent{\bf Transition matrix computation:}
We encode each annotated frame $I_{a_i}\in\calI_a$ into $\bfF_{a_i}$ to obtain the annotated frame feature set $\calF_a=\{\bfF_{a_1}, \ldots, \bfF_{a_N}\}$. Each frame feature is an $HW \times C_1$ matrix, in which each row contains the $C_1$-dimensional feature vector for a pixel. Here, $H\times W$ is the spatial resolution of the feature. Then, using the $i$th annotated feature $\bfF_{a_i}$ and the target feature $\bfF_{t}$, we obtain the transition matrix
\begin{equation}
\bfA_{a_i\rightarrow t}=\textrm{softmax}\big(\phi_A(\bfF_t)\times \phi_A(\bfF_{a_i})^T\big)
\label{eq:Amatrix}
\end{equation}
of size $HW \times HW$. Here, $\phi_A$ is a feature transform, implemented by a learnable $1\times 1$ convolution, to reduce the dimension of each row vector from $C_1$ to $C_2$. Note in Figure~\ref{fig:OverallFlow} that a frame feature is used by different modules. To adapt the same feature for different purposes, we employ multiple feature transforms, including $\phi_A$.

In \eqref{eq:Amatrix}, the softmax operation is applied to each column. Thus, $\bfA_{a_i\rightarrow t}$ is a positive matrix with each column adding to 1. It is hence the transition matrix~\cite{LinearAlgebra}, whose entry in row $r$ and column $c$ represents the probability that the $c$th pixel in $I_{a_i}$ is mapped to the $r$th pixel in $I_t$. We compute the transition matrices from all annotated frames to $I_t$ to yield the  transition matrix set $\calA_{a\rightarrow t}=\{\bfA_{a_1\rightarrow t}, \ldots, \bfA_{a_N\rightarrow t}\}$.

\begin{figure}[t]
\centering
    \includegraphics[width=\linewidth]{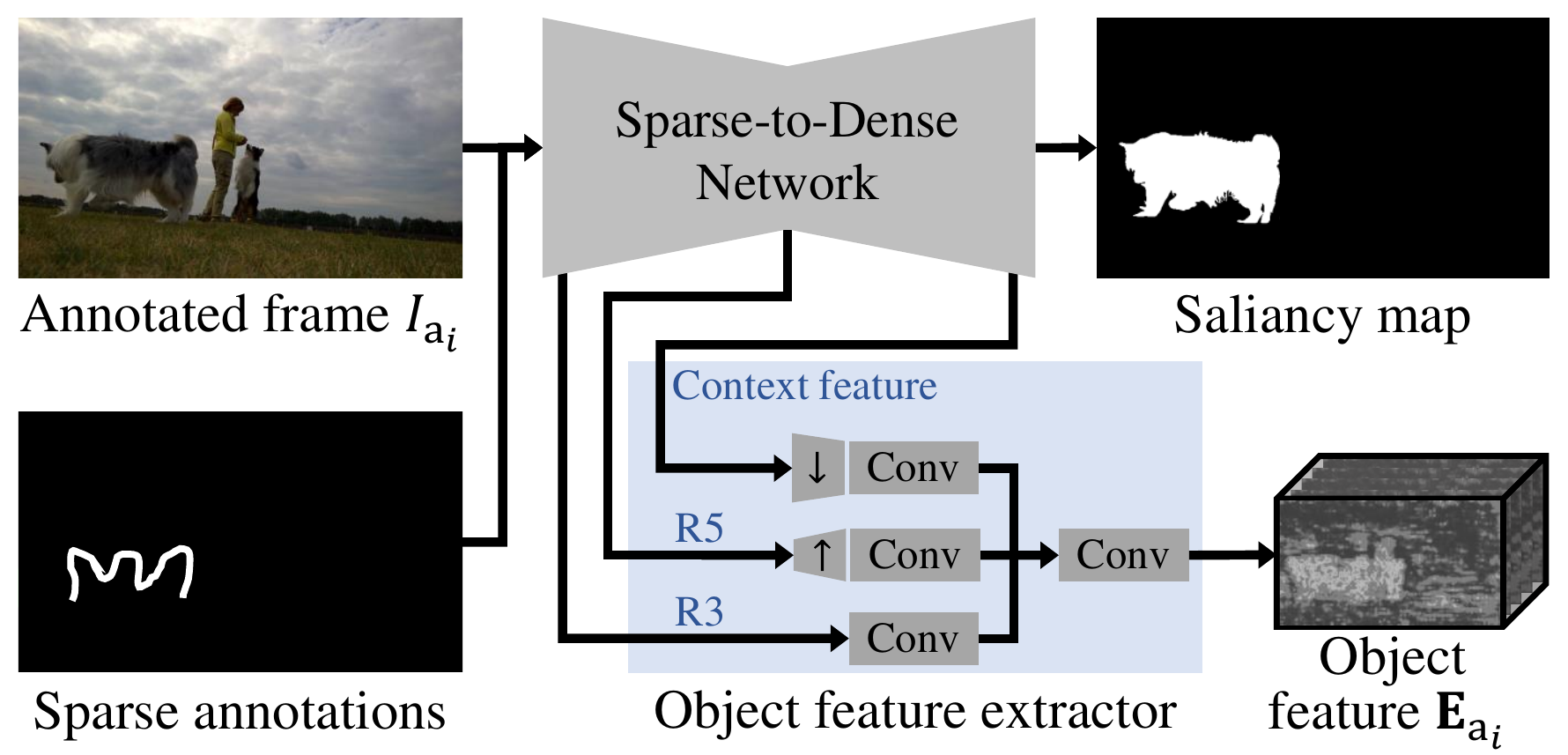}
\vspace{-0.5cm}
\caption{A diagram of the object feature extraction.}
\label{fig:ObjectFE}
\vspace{-0.1cm}
\end{figure}

\vspace*{0.15cm}
\noindent{\bf Object feature transfer:} We use annotations on $I_{a_i}$ to generate an object saliency map via a sparse-to-dense network in Figure~\ref{fig:ObjectFE}. We adopt A-Net \cite{Heo2020ECCV}, the encoder of which is based on SE-ResNet50~\cite{hu2018squeeze}, as the sparse-to-dense network. To form the feature of the query object, we combine three intermediate features: R3 and R5 features from the encoder and the context feature of the penultimate layer of the decoder. After making their spatial resolutions identical, we convolve and then concatenate them. Then, the concatenated feature passes through another convolution layer to form the object feature $\bfE_{a_i}$ of size $HW \times C_3$. Consequently, the object feature set $\calE_a=\{\bfE_{a_1}, \ldots, \bfE_{a_N}\}$ is obtained from the $N$ annotated frames.

We transfer all object features in $\calE_a$ to the target frame $I_t$ by
\begin{equation}
\bfE_{t|a_i}= \bfA_{a_i\rightarrow t}\times \bfE_{a_i}
\label{eq:Otransfer}
\end{equation}
using the transition matrix $\bfA_{a_i\rightarrow t}$ in \eqref{eq:Amatrix}. Thus, we have the transferred object feature set ${\calE}_{t|a}=\{{\bfE}_{t|a_1}, \ldots, {\bfE}_{t|a_N}\}$, which encodes the object information in $I_t$ approximately. Since the reliability of the transition matrix $\bfA_{a_i\rightarrow t}$ is different for each $i$, it is unreasonable to exploit $\bfE_{t|a_i}$ equally for the query object segmentation. To address this issue, we propose the R-attention mechanism.

\begin{figure}[t]
\centering
    \includegraphics[width=\linewidth]{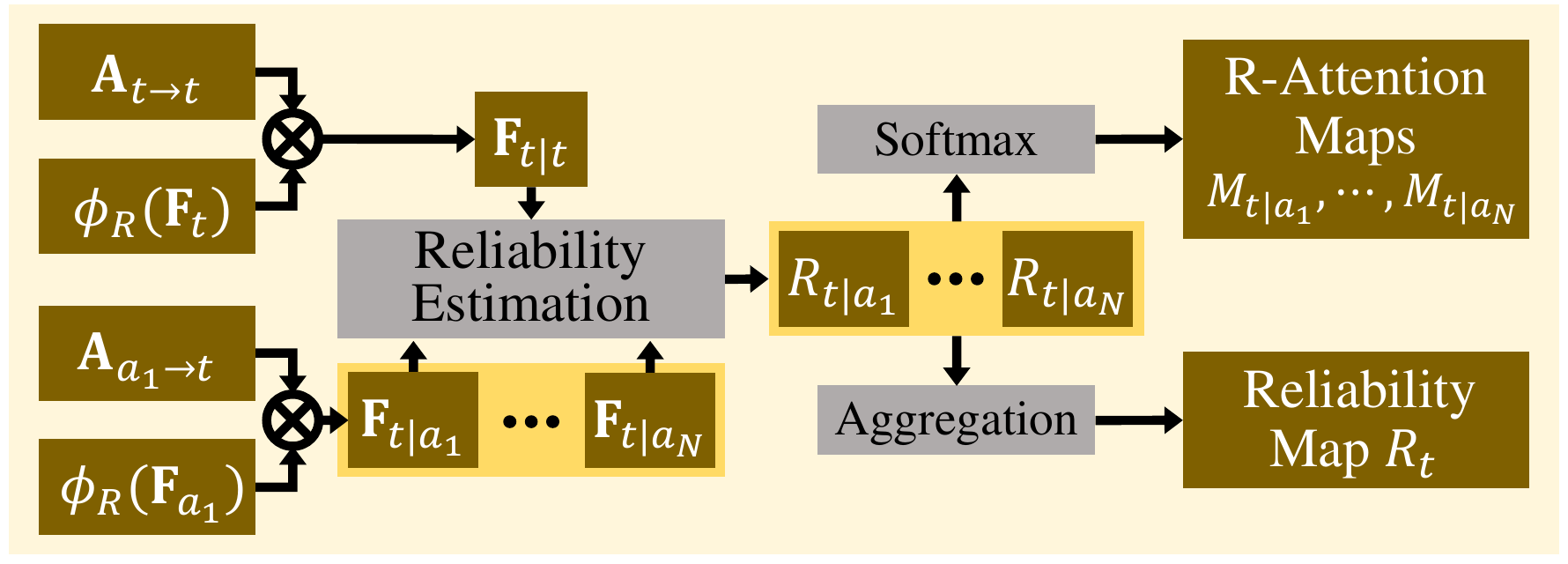}
\vspace{-0.6cm}
\caption{A diagram of the R-attention module.}
\label{fig:FlowRAtt}
\vspace{-0.1cm}
\end{figure}

\vspace*{0.15cm}
\noindent{\bf R-attention:} Figure~\ref{fig:FlowRAtt} shows how to generate R-attention maps. Similar to \eqref{eq:Otransfer}, let ${\bfF}_{t_2|t_1}=\bfA_{t_1\rightarrow t_2}\times \phi_R(\bfF_{t_1})$ denote the transferred frame feature from $I_{t_1}$ to $I_{t_2}$, where $\phi_R$ is a feature transform. We obtain the feature difference matrix
\begin{equation}
\bfD_{t|a_i} = [{\bfF}_{t|a_i} - {\bfF}_{t|t}]^{\circ2}
\end{equation}
where $\circ2$ is the entry-wise power operator. Note that its entry $\bfD_{t|a_i}(p,c)$ equals the squared distance between the $c$th feature components of pixel $p$ in ${\bfF}_{t|a_i}$ and ${\bfF}_{t|t}$. Ideally, all entries in $\bfD_{t|a_i}$ should be near zero, because both ${\bfF}_{t|a_i}$ and ${\bfF}_{t|t}$ represent the same frame $I_t$. However, they are not in practice, since the transition matrix $\bfA_{a_i\rightarrow t}$ in \eqref{eq:Otransfer} is imperfect. For the same reason, the transferred object feature $\bfE_{t|a_i}$ in \eqref{eq:Otransfer} may be unreliable. Hence, we define the reliability map $R_{t|a_i}$ for $\bfE_{t|a_i}$ as
\begin{equation}
R_{t|a_i}(p)= \frac{1}{\max_c \bfD_{t|a_i}(p,c)+\epsilon} \quad \text{ for each } p,
\label{eq:reliability}
\vspace{-0.1cm}
\end{equation}
where $\epsilon$ is a small positive number to prevent division by zero. A large value of $R_{t|a_i}(p)$ indicates that the $p$th row vector (\ie feature vector for pixel $p$) in $\bfE_{t|a_i}$ is reliable.

By applying the softmax function over the $N$ reliability maps, we generate the R-attention map $M_{t|a_i}$ for the transferred object feature  $\bfE_{t|a_i}$, which is given by
\begin{equation}
M_{t|a_i}(p)= \frac{\exp R_{t|a_i}(p)}{\sum_{k=1}^{N}\exp  R_{t|a_k}(p)} \quad \text{ for each } p.
\label{eq:Rattention}
\end{equation}
Next, we obtain the interfused object feature $\bfG_t$ by fusing all transferred object features using the R-attention maps,
\begin{equation}
\bfG_t = \sum_{i=1}^{N} M_{t|a_i} \otimes {\bfE}_{t|a_i}
\label{eq:interfused}
\end{equation}
where $\otimes$ means that $M_{t|a_i}$ is multiplied entry-wise to each column in $\bfE_{t|a_i}$. Through this R-attention mechanism, the interfused feature $\bfG_t$ contains more reliable information about the query object than each individual ${\bfE}_{t|a_i}$ does.

Furthermore, by aggregating $R_{t|a_1}, \ldots, R_{t|a_N}$, we generate the overall reliability map $R_t$ by
\begin{equation}
R_t(p)= \max_{i}\exp\left(R_{t|a_i}(p)-{\textstyle \small \frac{1}{\epsilon}}\right).
\label{eq:OverallR}
\end{equation}
Note from \eqref{eq:reliability} and \eqref{eq:OverallR} that each $R_t(p)$ is in the range $[0, 1]$, with 1 indicating the maximum reliability level. A high $R_t(p)$ means that the interfused feature vector for pixel $p$ in $\bfG_t$ is reliable. Because $\bfG_t$ plays an essential role in segmenting the target frame $I_t$, $R_t(p)$ also represents the reliability of the segmentation result.  Section~\ref{subsec:inference} describes how to use $R_t$ to guide the interactive VOS process.

\subsection{Overlapped Object Feature} \label{subsec:diffusion}

\begin{figure}[t]
\centering
    \includegraphics[width=\linewidth]{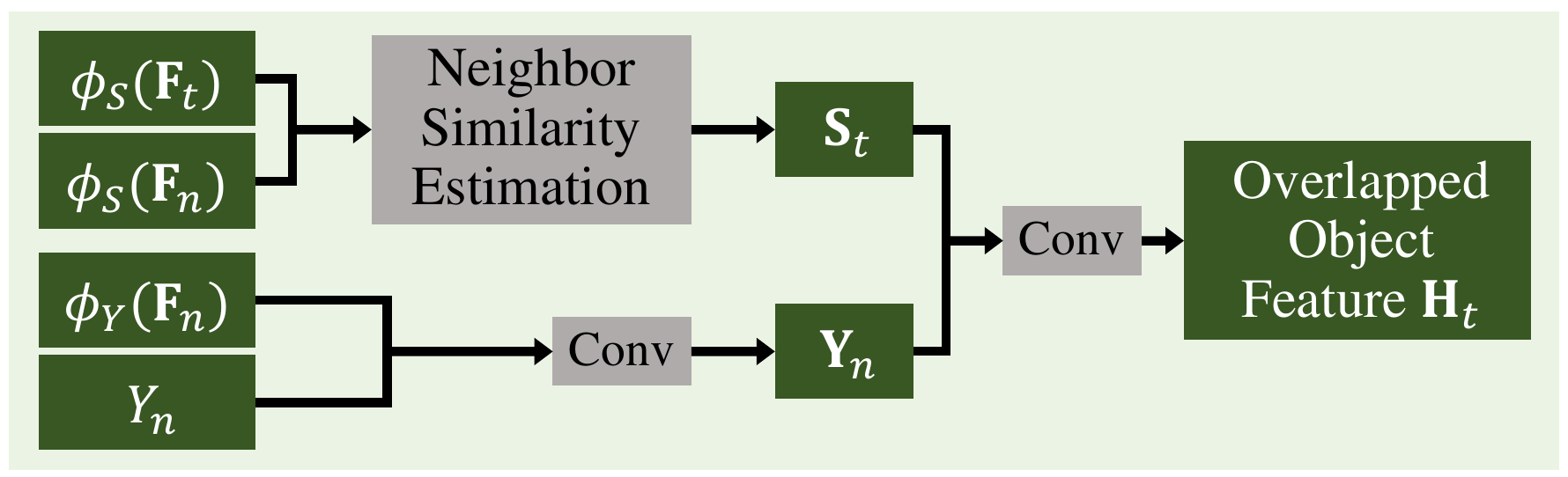}
\vspace{-0.6cm}
\caption{A diagram of the intersection-aware propagation.}
\label{fig:FlowInter}
\vspace{-0.1cm}
\end{figure}

\begin{figure}[t]
\centering
    \includegraphics[width=\linewidth]{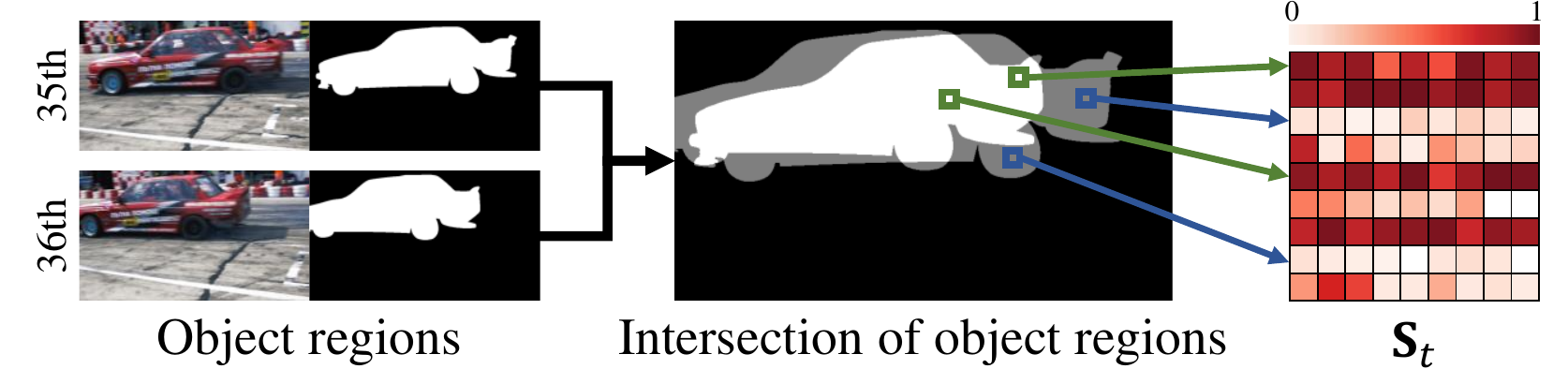}
    \vspace{-0.5cm}
\caption{An example of the neighbor similarity $\bfS_t$. The green pixels belong to the intersection, whereas the blue ones do not.}
\label{fig:InterSimFeature}
\vspace{-0.1cm}
\end{figure}

When segmenting the target frame $I_t$, the segmentation result $Y_n$ of the neighbor frame $I_n\in\{I_{t-1}, I_{t+1}\}$ is available according to the segmentation direction. We exploit this neighbor information to delineate the query object in $I_t$ more accurately. As shown in Figure~\ref{fig:FlowInter}, we first obtain the neighbor similarity
\begin{equation}
\bfS_t = \exp\big( -[\phi_S(\bfF_t) - \phi_S(\bfF_n)]^{\circ2} \big)
\end{equation}
using a feature transform $\phi_S$. $\bfS_t$ represents how similar the features of $I_t$ and $I_n$ are to each other. A row vector in $\bfS_t$ tends to have entries near one, when the corresponding pixel belongs to the intersection of the same object between the adjacent frames, as illustrated in Figure~\ref{fig:InterSimFeature}.

Next, we convert $\bfF_n$ via another feature transform $\phi_Y$ and concatenate it with $Y_n$. We convolve this concatenated signal to yield $\bfY_n$ that represents the query object in $I_n$. We then concatenate $\bfS_t$ and $\bfY_n$ and use another convolution layer to obtain the overlapped object feature $\bfH_t$. Note that $\bfS_t$ indicates the overlapped region (or intersection) of the query object in $I_n$ and $I_t$. Therefore, we combine $\bfS_t$ and $\bfY_n$ to recognize the query object feature in the overlapped region, and hand over this information to the target frame $I_t$ via $\bfH_t$. This intersection-aware propagation of object information enables the segment head to exploit the neighbor information selectively and reliably.

Last, we use the frame feature $\bfF_t$, the interfused object feature $\bfG_t$, and the overlapped object feature $\bfH_t$ to obtain the segmentation mask $Y_t$.

\subsection{Guided Interactive VOS Process}
\label{subsec:guide}
\begin{figure}[t]
\centering
    \includegraphics[width=\linewidth]{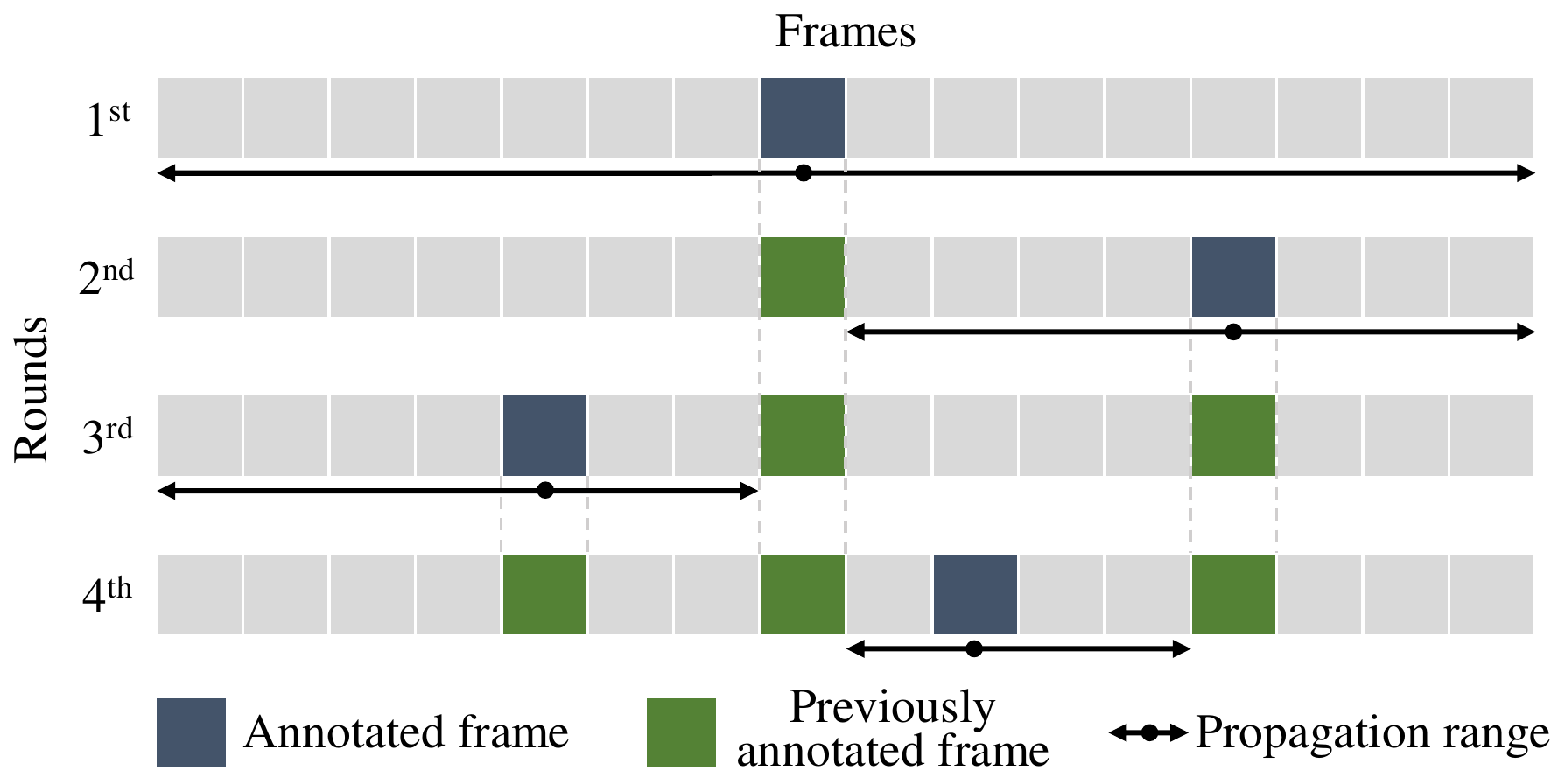}
    \vspace{-0.5cm}
\caption{Illustration of the interactive VOS process.}
\label{fig:Inference}
\vspace{-0.1cm}
\end{figure}

In the first round, to segment the target frame $I_{t} = I_{a_1}$, we set $I_{n}=I_{a_1}$ and substitute $Y_n$ with the saliency map of $I_t$ generated by the sparse-to-dense network. This is because there is no neighbor frame already segmented. After performing the segmentation of $I_{a_1}$, we propagate the segmentation mask $Y_{a_1}$ bidirectionally to segment the other frames. In the second round, sparse annotations are given on $I_{a_2}$ to refine its segmentation result ${Y}_{a_2}$, which is also propagated bidirectionally until another annotated frame is met. This is repeated until the user is satisfied with the VOS result. Figure~\ref{fig:Inference} illustrates this process.

Suppose that there are $K$ query objects with their annotations. Then, for each query object in each target frame $I_t$, the segmentation head uses the three features $\bfF_t$, $\bfG_t$, $\bfH_t$ to estimate the object probability map. Consequently, we have $\hat{Y}_{t,1}, \ldots, \hat{Y}_{t,K}$, where $\hat{Y}_{t, k}$ denotes the probability map for the $k$th query object. By applying the soft aggregation scheme~\cite{Oh2019ICCV} to these maps and then allocating each pixel to the background or the query object with the highest probability, we yield the binary segmentation masks $Y_{t, 1}, \ldots, Y_{t,K}$ at the target frame.

In interactive VOS, it is important to enable the user to provide annotations quickly with less effort. Therefore, after performing the segmentation in each round, we compute the R-score $r_t$ of each frame $I_t$,
\begin{equation}
r_t= \frac{\alpha}{\|\calU_t \|}\sum_{p\in\calU}^{}R_t(p) + \frac{1-\alpha}{\|\calO_t \|}\sum_{p\in\calO}^{}R_t(p)
\label{eq:Rscore}
\end{equation}
where $\calU_t$ is the set of pixels in the entire frame and $\calO_t$ is the union set of pixels in the segmented object regions. When $\alpha = 0$, the pixel-wise reliability in \eqref{eq:OverallR} is averaged over the foreground segments only. When $\alpha=1$, it is averaged over the entire frame. In this work, $\alpha$ is set to 0.5 to consider all pixels but also to emphasize the foreground segments.

\vspace*{0.15cm}
\noindent{\bf Guided selection of annotated frames:}
In practice, it takes considerable time to find the most poorly segmented frame and provide annotations. To alleviate this problem, from the second round, the R-scores $\{r_1, ..., r_T\}$ are used to guide the user to provide additional annotations for the next round. Instead of a time-consuming search over the entire video, the user can select the frames for annotations in two ways.

\begin{itemize}[leftmargin=*]
\vspace*{-0.15cm}
\item {\bf RS1:} The single frame with the lowest R-score is chosen for next annotations.
\vspace*{-0.15cm}
\item {\bf RS4:} The four frames with the lowest R-scores are determined subject to the constraint that their time distances are at least $T/10$. In the interactive VOS simulation, the most poorly segmented frame is selected among the four frames, by comparing the segmentation results with the ground-truth. In real applications, users are provided with the segmentation results of these four guided frames only. Then, the user chooses a frame among them and provides annotations. The interactive process terminates, when the user is satisfied with the guided frames.
\end{itemize}
\vspace*{-0.2cm}

\subsection{Implementation Details} \label{subsec:implement}

\noindent{\bf Network details:}
To encode each frame $I_t$ to the frame feature $\bfF_t$, we employ SE-ResNet50~\cite{hu2018squeeze} from the first layer to R4 with an output stride 8. Thus, $H$ and $W$ in Section~\ref{subsec:affinity} are $\frac{1}{8}$ of the height and width of an input frame, respectively. The dimension $C_1$ of each frame feature vector is 1,024, while the dimensions $C_2$ of output vectors of the four feature transforms $\phi_A$, $\phi_R$, $\phi_S$, and $\phi_Y$ are equally set to 128. Also, $C_3$ for $\bfE_{t}$, $\bfG_{t}$, and $\bfH_{t}$ is set to 256. For the segmentation head, we adopt the decoder architecture in \cite{Heo2020ECCV}.

\begin{table}[t]\footnotesize\centering\renewcommand\arraystretch{1.15}
\caption{Comparative assessment of the proposed algorithm with the state-of-the-art interactive VOS algorithms on the DAVIS2017 validation set. The best results are boldfaced.}
\vspace*{-0.2cm}
\begin{tabular}[t]{+L{2.2cm}^C{0.8cm}^C{0.8cm}^C{1.2cm}^C{1.15cm}} \toprule
                                        & AUC-J      & J@60s      & AUC-J\&F   & J\&F@60s\\ \midrule
Oh~\etal~\cite{Oh2019CVPR}              & 0.691      & 0.734      & 0.778   & 0.787\\
Miao~\etal~\cite{Miao2020CVPR}          & 0.749      & 0.761      & 0.787      & 0.795\\
Heo~\etal~\cite{Heo2020ECCV}            & 0.771      & 0.790      & 0.809      & 0.827\\
Oh~\etal~\cite{Oh2020TPAMI}             &    -       &    -       & 0.839   & 0.848\\\midrule
Proposed-GT                                & 0.817 & 0.826 & 0.853 & 0.863\\
Proposed-RS1                            & 0.818 & 0.827 & 0.855 & 0.864\\
Proposed-RS4                            & \bf{0.820} & \bf{0.829} & \bf{0.856} & \bf{0.866}\\ \bottomrule

\end{tabular}
\label{tb:ComparativeTab}
\vspace{-0.3cm}.
\end{table}

\vspace*{0.15cm}
\noindent{\bf Training:}
We use the training sets of DAVIS2017~\cite{DAVIS2017} and YouTube-VOS~\cite{Youtube2018}. To emulate the first round, we randomly form a mini-sequence by taking five consecutive frames (one annotated frame and four target frames) from a video sequence. To emulate the second round, we pick one additional frame as the second annotated frame. In the training, we proceed up to the second round due to limited GPU memories. To imitate sparse annotations, we use two types: 1) random points and 2) scribble generation in~\cite{DAVISchallenge2018}. More implementation details are in the supplementary document.

\section{Experimental Results} \label{sec4:Experiments}
First, we compare the proposed GIS algorithm with the state-of-the-art interactive VOS algorithms. Second, we analyze the proposed algorithm through various ablation studies and visualization of feature maps. Third, we perform a user study to demonstrate the effectiveness of the proposed algorithm in real applications.

\subsection{Comparative Assessment} \label{subsec:compareassess}
Interactive VOS simulation is conducted on two datasets: DAVIS2017~\cite{DAVIS2017} and YouTube-IVOS.

\vspace*{0.15cm}
\noindent{\bf DAVIS2017:}
In the DAVIS interactive VOS simulation~\cite{DAVISchallenge2018}, human interactions are emulated up to 8 rounds by an algorithm. In each round, after VOS is performed, the algorithm determines the frame with the poorest performance, by comparing the segmentation results with the ground-truth, and provides additional annotations on it. We follow this procedure for the comparison with conventional algorithms. We also follow the two guided procedures RS1 and RS4 in Section~\ref{subsec:guide} to confirm the effectiveness of R-scores. The validation set of 30 video sequences in DAVIS2017 is used for the assessment. For each video sequence, three distinct initial scribbles are provided, which means that the performance is averaged over 90 interactive VOS trials.

\begin{figure}[t]
\centering
    \includegraphics[width=1.0\linewidth]{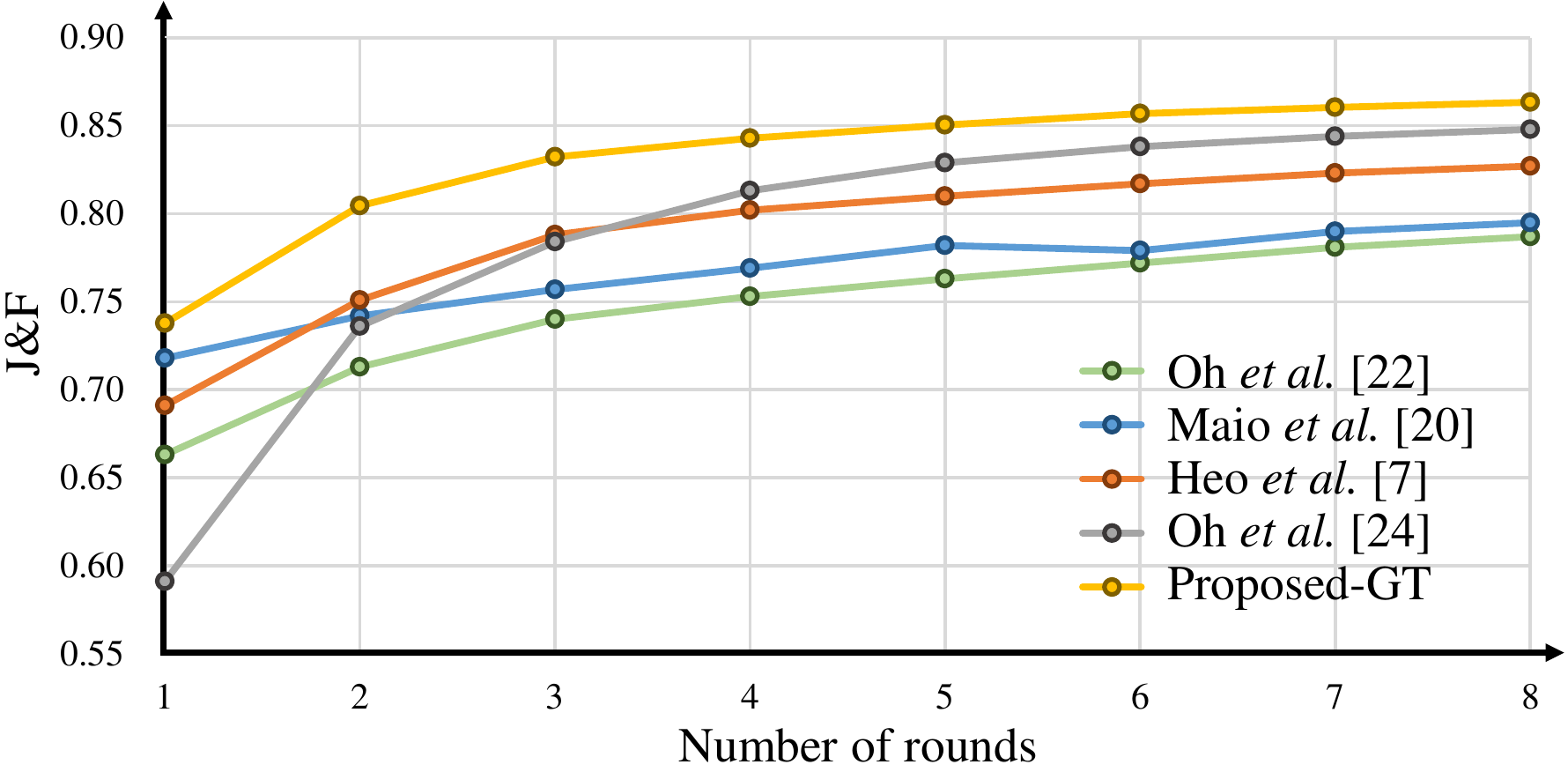}
\vspace{-0.5cm}
\caption{Comparison of J\&F scores on the DAVIS2017 validation set according to the rounds.}
\label{fig:ComparativeGraph}
\vspace{-0.4cm}
\end{figure}

We quantify the segmentation performance using the region similarity (J) and the contour accuracy (F). We measure the area under the curve (AUC) of a performance-versus-time graph from 0 to 488 seconds for J score (AUC-J) or for joint J and F scores (AUC-J\&F). Also, we measure the performance at 60 seconds for J score (J@60s) or for joint J and F scores (J\&F@60s) to assess how accurately the segmentation is carried out within 60 seconds.

Table~\ref{tb:ComparativeTab} compares the proposed GIS algorithm with the recent state-of-the-art algorithms~\cite{Oh2019CVPR,Miao2020CVPR,Heo2020ECCV,Oh2020TPAMI}, in which Proposed-GT, Proposed-RS1, and Proposed-RS4 denote the settings when the ground-truth, RS1, and RS4 are used to choose frames to be annotated in next rounds, respectively.
Note that Proposed-GT has the same experimental conditions as the conventional algorithms~\cite{Oh2019CVPR,Miao2020CVPR,Heo2020ECCV,Oh2020TPAMI}.
In Proposed-RS4, the DAVIS algorithm selects
the poorest frame among the four guided frames using the ground-truth. In all settings, the DAVIS algorithm provides annotations. The scores of the conventional algorithms are provided by the respective authors. It can be observed from Table~\ref{tb:ComparativeTab} that the proposed algorithm outperforms the state-of-the-arts by significant margins in all metrics. In other words, the proposed algorithm performs the best in both accuracy and speed. Also, Proposed-RS1 and Proposed-RS4 perform better than Proposed-GT, by employing R-scores and selecting annotated frames effectively. Figure~\ref{fig:ComparativeGraph} shows the J\&F scores according to the rounds. The proposed algorithm yields the best score in every round with no exception.

\vspace*{0.15cm}
\noindent{\bf YouTube-IVOS:}
For extensive experiments, we construct the YouTube-IVOS dataset from YouTube-VOS~\cite{Youtube2018}, which is the largest VOS dataset. For the DAVIS algorithm to emulate user interactions, ground-truth segmentation masks are needed. Since the validation set in YouTube-VOS does not provide the ground-truth, we sample 200 videos from its training set to compose YouTube-IVOS. For each video, we generate four different initial annotations by varying the number of point clicks. Specifically, we randomly pick 5, 10, 20, and 50 point clicks from the ground-truth mask for each query object and then use those clicks as annotations in the first round. The interactive VOS is performed up to 4 rounds, since the performance is saturated in early rounds due to the short lengths of the YouTube-VOS videos.

Table~\ref{tb:YoutubeIVOS} compares the average J\&F scores of the proposed algorithm with those of Miao \etal \cite{Miao2020CVPR} and Heo \etal \cite{Heo2020ECCV} according to the rounds. Notice that \cite{Oh2019CVPR} and \cite{Oh2020TPAMI} are not compared in this test, because their full source codes are unavailable. In this test, the proposed GIS network and the Heo \etal's network are trained without the 200 videos in YouTube-IVOS. We see that the proposed algorithm outperforms the other algorithms meaningfully in all rounds.

\begin{table}[t]\footnotesize\centering\renewcommand\arraystretch{1.15}
\caption{Comparative assessment of the proposed algorithm with the state-of-the-art interactive VOS algorithms on the Youtube-IVOS dataset.}
\vspace*{-0.2cm}
\begin{tabular}[t]{L{2.0cm}^C{1.0cm}^C{1.0cm}^C{1.0cm}^C{1.0cm}}
\toprule
                                & J\&F-1st   & J\&F-2nd    & J\&F-3rd   & J\&F-4th    \\
\midrule
Miao~\etal~\cite{Miao2020CVPR}  & 0.525      & 0.620       & 0.674      & 0.706 \\
Heo~\etal~\cite{Heo2020ECCV}    & 0.643      & 0.721       & 0.768      & 0.797 \\
Proposed-GT                     & \bf{0.672}      & \bf{0.754}       & \bf{0.806}      & \bf{0.830}\\
\bottomrule
\end{tabular}
\label{tb:YoutubeIVOS}
\vspace{-0.2cm}
\end{table}

\begin{figure}[t]
\vspace*{0.3cm}
\centering
    \includegraphics[width=\linewidth]{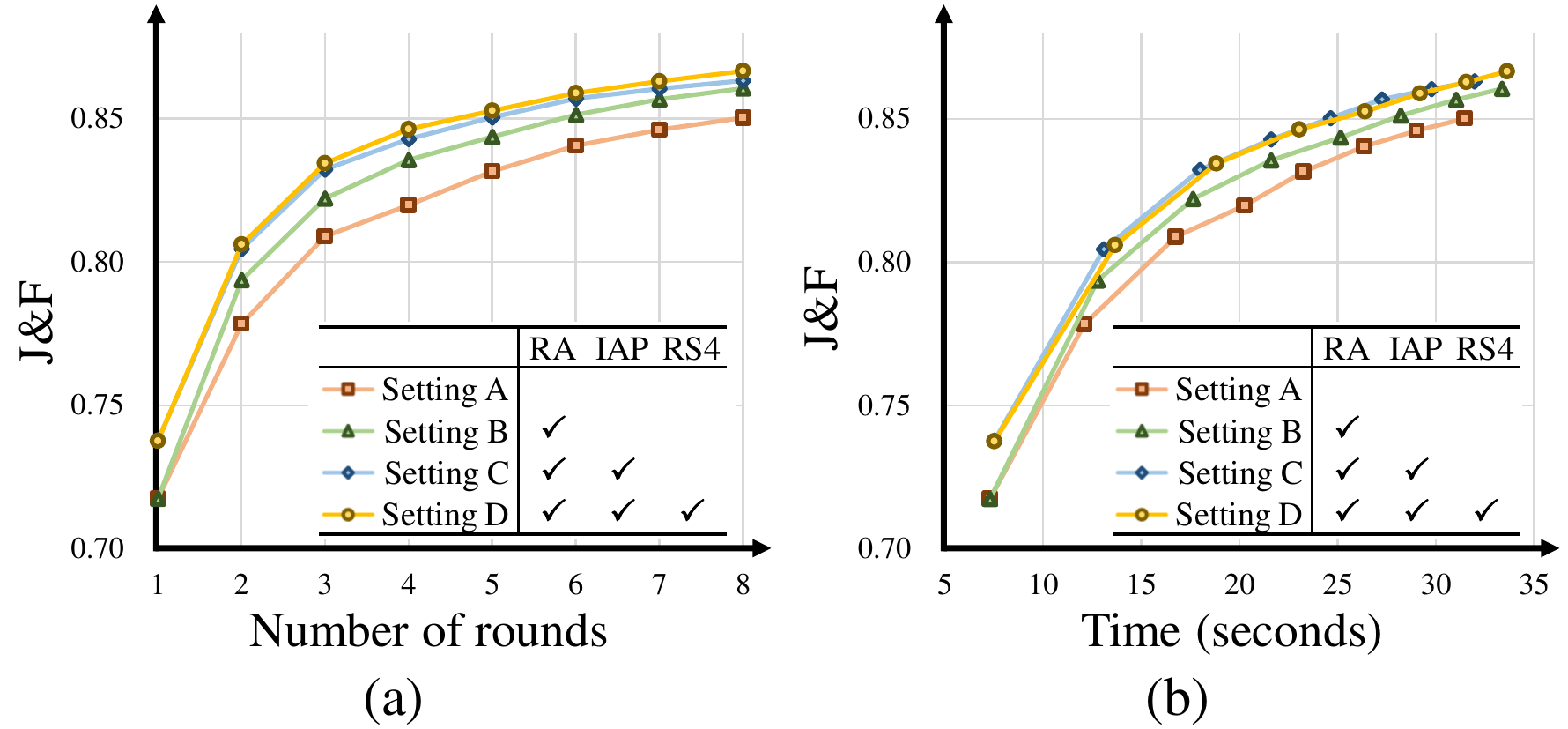}
\caption{Ablation study results on the DAVIS2017 validation set.}
\label{fig:AblationResult}
\vspace{-0.2cm}
\end{figure}

\subsection{Analysis} \label{subsec:analysis}
\vspace*{-0.0cm}

\noindent{\bf Ablation study:}
We analyze the effectiveness of three components in the proposed algorithm:
\vspace*{-0.15cm}
\begin{enumerate}[label=(\arabic*)]
\itemsep0mm
\item R-attention (RA)
\item Intersection-aware propagation (IAP)
\item R-score guidance with four candidate frames (RS4)
\end{enumerate}
\vspace*{-0.15cm}
Figure~\ref{fig:AblationResult} plots the J\&F scores of four settings A, B, C, and D, combining these three components, on the DAVIS2017 validation set. First, using R-attention maps, the overall performance in every round increases significantly (gaps between A and B). Also, the intersection-aware propagation improves the performance especially in early rounds (gaps between B and C). The R-score guidance affects the performance only slightly (gaps between C and D), because in this test the computer searches the frames to be annotated using the ground-truth. In real applications, the R-score guidance enables human users to search the frames efficiently and thus reduces the overall segmentation time, as will be verified in later experiments.

\vspace*{0.15cm}
\noindent{\bf Intersection-aware propagation:}
We compare the proposed IAP module with two existing propagation methods: the local distance map (LDM) in \cite{Miao2020CVPR} and the local transfer module (LTM) in \cite{Heo2020ECCV}. LDM matches each pixel in a target frame to a local region in a neighbor frame at the feature level to estimate a local distance map, while LTM transfers the segmentation result of a neighbor frame based on the local affinity. In Table~\ref{tb:IFAnal}, the baseline means the proposed network without IAP. In other words, the baseline uses only the frame feature $\bfF_t$ and the interfused object feature $\bfG_t$ for the segmentation. We plug LDM or LTM into the baseline. Table~\ref{tb:IFAnal} compares the J\&F scores in 1st, 3rd, and 5th rounds and the segmentation speeds (frames per second, FPS). Compared with LDM and LTM, the proposed IAP helps the baseline network to achieve higher segmentation accuracies and a faster speed.

\begin{table}[t]\footnotesize\centering\renewcommand\arraystretch{1.15}
\caption{Comparison of the proposed intersection-aware propagation (IAP) module with conventional local propagation methods on the DAVIS2017 validation dataset.}
\vspace*{-0.2cm}
\begin{tabular}[t]{L{2.3cm} C{1.0cm} C{1.0cm} C{1.0cm} C{0.8cm}} \toprule
                                      & J\&F-1st & J\&F-3rd & J\&F-5th & FPS  \\ \midrule
Baseline                              & 0.717    & 0.821    & 0.839    & 9.26 \\ \midrule
Baseline+LDM~\cite{Miao2020CVPR}      & 0.727    & 0.824    & 0.844    & 8.23 \\
Baseline+LTM~\cite{Heo2020ECCV}       & 0.730    & 0.828    & 0.846    & 8.69 \\
Baseline+IAP                          & \bf{0.737}    & \bf{0.832}    & \bf{0.850}    & \bf{8.72} \\  \bottomrule
\end{tabular}
\label{tb:IFAnal}
\vspace{-0.1cm}
\end{table}

\begin{figure*}[t]
\centering
    \includegraphics[width=0.95\linewidth]{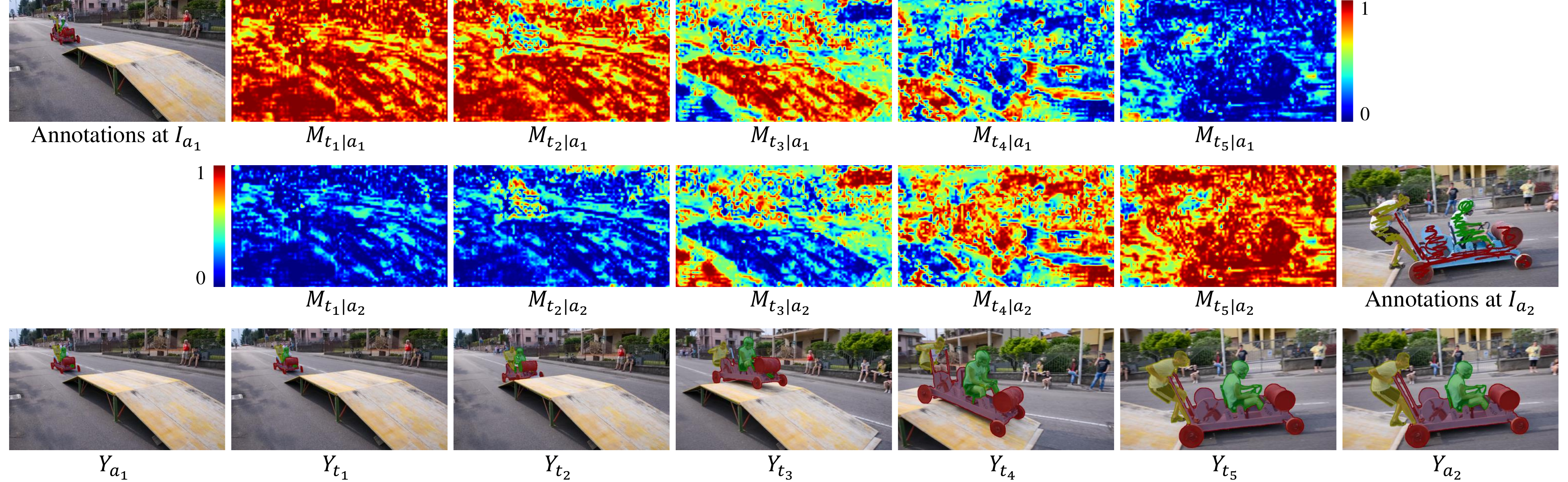}
\vspace{-0.1cm}
\caption{Visualization of R-attention maps and segmentation results on the ``soapbox'' sequence in the second round.}
\label{fig:AttMap}
\vspace{-0.3cm}
\end{figure*}

 \begin{figure}[t]
\centering
    \includegraphics[width=\linewidth]{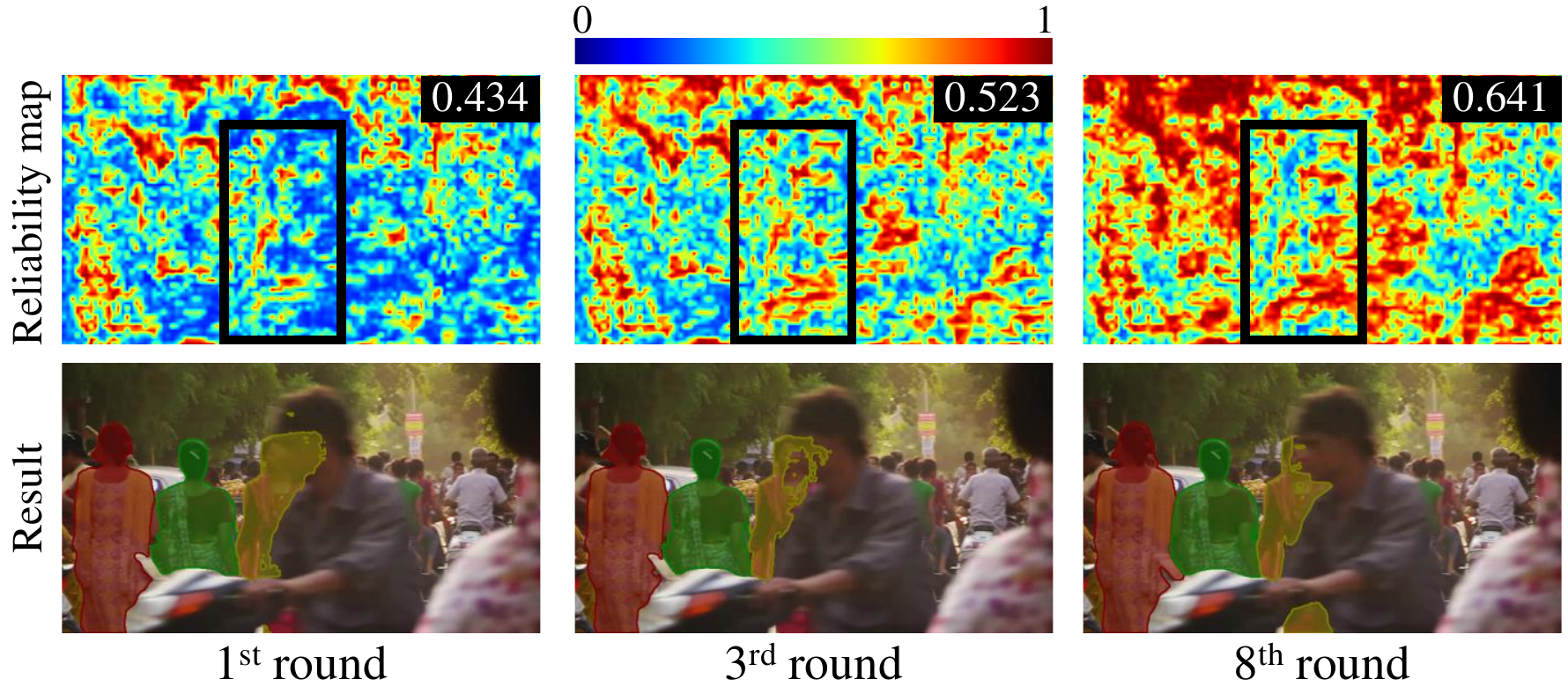}
\caption{The reliability maps and R-scores for the 33rd frame of the ``India'' video in the first, third, and eighth rounds. The number in the upper right corner of each reliability map is the R-score.}
\label{fig:RelMap}
\vspace{-0.25cm}
\end{figure}
\vspace*{0.15cm}
\noindent{\bf R-attention maps:}
Figure~\ref{fig:AttMap} illustrates R-attention maps $M_{t|a_i}$ in \eqref{eq:Rattention}.  Specifically, the first and second rows in Figure~\ref{fig:AttMap} show the R-attention maps of the first annotation at $I_{a_1}$ and the second annotation at $I_{a_2}$, respectively. We sample time instances $t_1 \sim t_5$ uniformly between $a_1$ and $a_2$. Note that R-attention values depend on the pixel-wise feature similarities between annotated and target frames. For instance, $M_{t_4|a_1}$ has low R-attention values on the query objects (cart and two people), which have significantly different appearance and sizes between $I_{a_1}$ and $I_{t_4}$. Also, $M_{t_3|a_2}$ has low R-attention values around the plywood ramp that almost disappears in $I_{a_2}$. This example indicates that R-attention module faithfully provides the reliability information of the transferred object feature from the annotated frame to the target frame.

\vspace*{0.15cm}
\noindent{\bf Reliability maps and R-scores:}
Examples of reliability maps and R-scores are in Figure~\ref{fig:RelMap}. In the first round, a poor segmentation result for a woman marked in yellow is obtained, because she is partly occluded by a man riding a bike. Thus, the proposed algorithm yields the reliability map that has low values within the black box containing the woman. As the segmentation result is refined in subsequent rounds, the reliability map has higher values and the R-score gets larger. This means that both the reliability map $R_t$ in \eqref{eq:reliability} and the R-score $r_t$ in \eqref{eq:Rscore} are good indicators of how accurately the target frame $I_t$ is segmented.

\subsection{User Study} \label{subsec:userstudy}
We conducted a user study to assess the proposed GIS algorithm in real applications. We recruited 12 volunteers, who provided scribbles iteratively until they were satisfied. We measured the average time in seconds per video (SPV), including the time for providing scribbles, the running time of the algorithm, and the time for finding unsatisfactory frames. Also, we measured the average rounds per video (RPV) and the average J\&F score over all video sequences.

\begin{figure}[t]
\centering
    \includegraphics[width=1\linewidth]{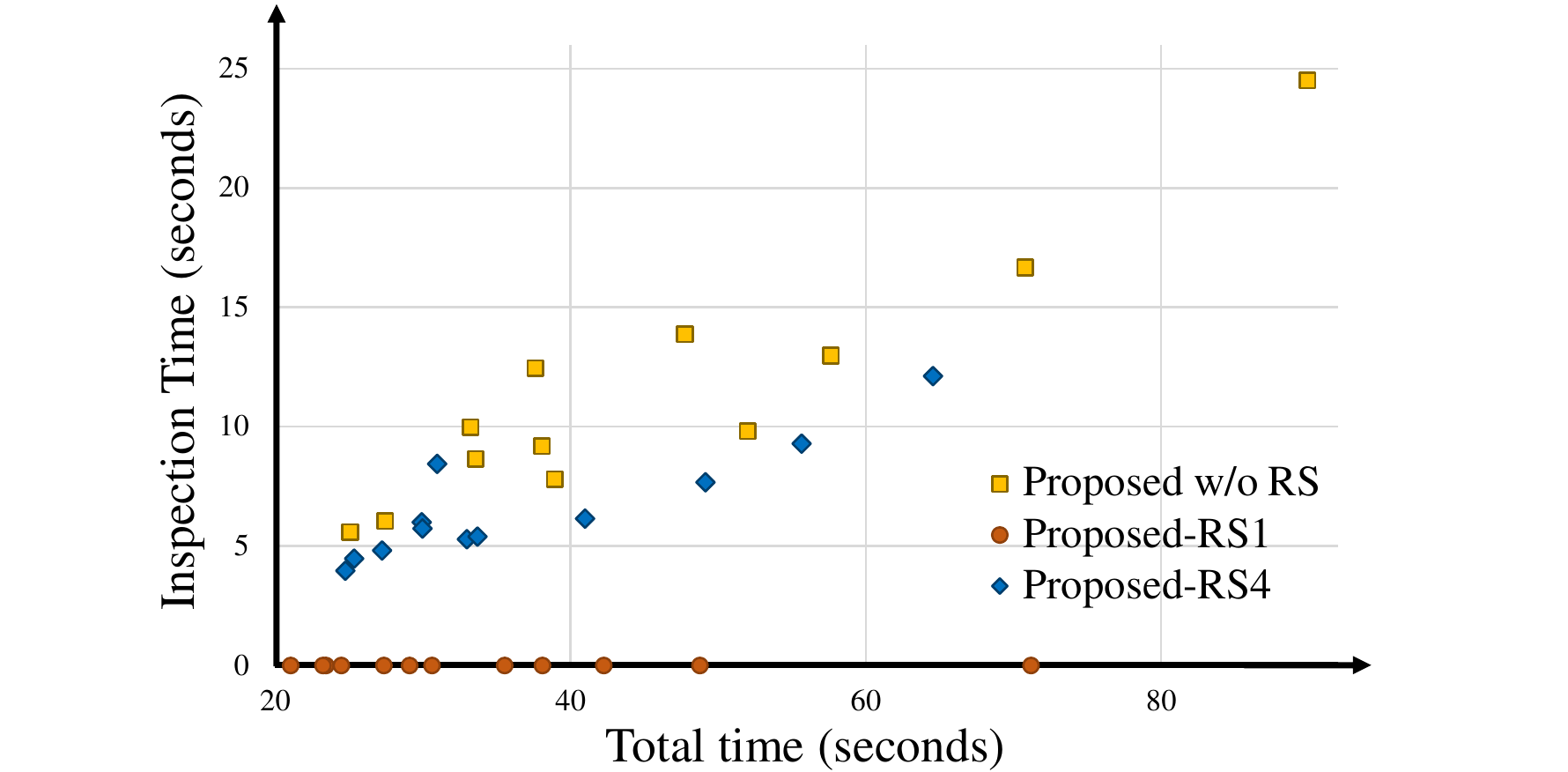}
\vspace{-0.1cm}
\caption{The total time and the inspection time for selecting unsatisfactory frames per video. Each user is represented by each mark.}
\label{fig:FvsT}
\vspace{-0.4cm}
\end{figure}

Table~\ref{tb:UserStudy} compares these user study results on the validation set in DAVIS2017~\cite{DAVIS2017}. `Proposed w/o RS' denotes the proposed algorithm without using R-scores. All three settings of the proposed algorithm outperform the state-of-the-art algorithm~\cite{Heo2020ECCV} in all metrics. This indicates that the proposed algorithm requires less running time and less interaction, while providing better segmentation results. Proposed-RS1 and Proposed-RS4 require less time to complete the process than `Proposed w/o RS' with only negligible performance degradation, by removing or reducing the time for selecting unsatisfactory frames as shown in Figure~\ref{fig:FvsT}. Especially, the total time for segmenting a video is significantly reduced in Proposed-RS1, which needs no time for inspection.

\section{Conclusions} \label{sec5:conclusion}
\vspace{-0.1cm}
We proposed the novel GIS algorithm for video objects based on R-attention and intersection-aware propagation. First, the interfused object feature is extracted by transferring query object information from annotated frames to a target frame using the R-attention module. Second, the overlapped object feature is obtained via the intersection-aware propagation using a neighbor frame. Then, the segmentation is performed using the frame feature, interfused object feature, and overlapped object feature. Moreover, we developed the GIS mechanism that enables users to determine next annotated frames quickly. Experimental results showed that the proposed algorithm outperforms the conventional algorithms significantly.

\begin{table}[t]\footnotesize\centering\renewcommand\arraystretch{1.15}
\caption{User study results.}
\vspace*{-0.25cm}
\begin{tabular}[t]{+L{2.1cm}^C{1.1cm}^C{0.9cm}^C{0.9cm}}
\toprule
                                  & SPV(s) & RPV           & J\&F  \\
\midrule
Heo~\etal~\cite{Heo2020ECCV}      & 66.83       & 2.42              & 0.769  \\
Proposed w/o RS                   & 46.01      & 1.89               & \bf{0.794}  \\
Proposed-RS1                      & \bf{34.59}      & 1.82             & 0.789  \\
Proposed-RS4                      & 37.10 & \bf{1.69}    & \bf{0.794}  \\
\bottomrule
\end{tabular}
\label{tb:UserStudy}
\end{table}

\section*{Acknowledgements}
\vspace{-0.1cm}
This work was supported in part by the National Research Foundation of Korea (NRF) grant funded by MSIT, Korea (No. NRF-2018R1A2B3003896), in part by
MSIT, Korea, under the ITRC support program (IITP-2020-2016-0-00464) supervised by the IITP,
and in part by the NRF grant funded by MSIT, Korea (No. NRF-2019R1F1A1062907).

{\small
\bibliographystyle{ieee}
\bibliography{IVOS_4551}
}
\end{document}